\documentclass{article}

\usepackage[numbers]{natbib}

\usepackage{amsmath,amsfonts,bm, amssymb}

\def\eqref#1{equation~\ref{#1}}

\def\1{\bm{1}}

\newcommand{\test}{\mathcal{D_{\mathrm{test}}}}

\def\va{{\bm{a}}}

\def\vc{{\bm{c}}}

\def\ve{{\bm{e}}}

\def\vg{{\bm{g}}}

\def\vo{{\bm{o}}}
\def\vp{{\bm{p}}}

\def\vs{{\bm{s}}}

\def\vx{{\bm{x}}}

\def\vz{{\bm{z}}}

\DeclareMathAlphabet{\mathsfit}{\encodingdefault}{\sfdefault}{m}{sl}
\SetMathAlphabet{\mathsfit}{bold}{\encodingdefault}{\sfdefault}{bx}{n}

\def\gA{{\mathcal{A}}}
\def\gB{{\mathcal{B}}}

\def\gE{{\mathcal{E}}}

\def\gG{{\mathcal{G}}}

\def\gI{{\mathcal{I}}}

\def\gO{{\mathcal{O}}}

\def\gS{{\mathcal{S}}}

\def\gU{{\mathcal{U}}}
\def\gV{{\mathcal{V}}}

\def\gX{{\mathcal{X}}}

\def\gZ{{\mathcal{Z}}}

\def\sB{{\mathbb{B}}}

\def\sR{{\mathbb{R}}}

\def\sU{{\mathbb{U}}}
\def\sV{{\mathbb{V}}}
\def\sW{{\mathbb{W}}}

\newcommand{\E}{\mathbb{E}}

\DeclareMathOperator*{\argmax}{arg\,max}
\DeclareMathOperator*{\argmin}{arg\,min}

\usepackage[final]{neurips_2019}

\usepackage[utf8]{inputenc} 
\usepackage[T1]{fontenc}    
\usepackage{hyperref}       
\usepackage{url}            
\usepackage{booktabs}       
\usepackage{amsfonts}       
\usepackage{nicefrac}       
\usepackage{microtype}      

\usepackage{wrapfig}
\usepackage[pdftex]{graphicx}
\usepackage{caption}
\usepackage{subcaption}
\RequirePackage{algorithm}
\RequirePackage{algorithmic}
\RequirePackage{eso-pic} 
\RequirePackage{forloop}
\usepackage{multicol}
\usepackage[mathscr]{euscript}
\usepackage{wrapfig,lipsum,booktabs}
\usepackage{makecell}
\usepackage{enumitem,kantlipsum}
\usepackage[export]{adjustbox}

\newcommand{\change}[1]{{ #1}}

\title{Language as an Abstraction\\ for Hierarchical Deep Reinforcement Learning}

\author{%
  Yiding Jiang\thanks{ Work done as a part of the Goolge AI Residency program} , Shixiang Gu, Kevin Murphy, Chelsea Finn\\
  Google Research\\
  \texttt{\{ydjiang,shanegu,kpmurphy,chelseaf\}@google.com}
}

\begin{document}

\maketitle

\begin{abstract}
Solving complex, temporally-extended tasks is a long-standing problem in reinforcement learning (RL). We hypothesize that one critical element of solving such problems is the notion of \emph{compositionality}. With the ability to learn concepts and sub-skills that can be composed to solve longer tasks, i.e. hierarchical RL, we can acquire temporally-extended behaviors. However, acquiring effective yet general abstractions for hierarchical RL is remarkably challenging. In this paper, we propose to use language as the abstraction, as it provides unique compositional structure, enabling fast learning and combinatorial generalization, while retaining tremendous flexibility, making it suitable for a variety of problems. Our approach learns an instruction-following low-level policy and a high-level policy that can reuse abstractions across tasks, in essence, permitting agents to reason using structured language. To study compositional task learning, we introduce an open-source object interaction environment built using the MuJoCo physics engine and the CLEVR engine. We find that, using our approach, agents can learn to solve to diverse, temporally-extended tasks such as object sorting and multi-object rearrangement, including from raw pixel observations. Our analysis reveals that the compositional nature of language is critical for learning diverse sub-skills and systematically generalizing to new sub-skills in comparison to non-compositional abstractions that use the same supervision.\footnote{Code and videos of the environment, and experiments are at \hyperlink{https://sites.google.com/view/hal-demo}{https://sites.google.com/view/hal-demo}}

\end{abstract}

\vspace{-0.27cm}
\section{Introduction}
\label{intro}
\vspace{-0.1cm}

Deep reinforcement learning offers a promising framework for enabling agents to autonomously acquire complex skills, and has demonstrated impressive performance on continuous control problems~\citep{levine2016end,schulman2015high} and games such as Atari~\citep{dqn} and Go~\citep{alphago}. 
However, the ability to learn a variety of compositional, long-horizon skills while generalizing to novel concepts remain an open challenge.
Long-horizon tasks demand sophisticated exploration strategies and structured reasoning, while generalization requires suitable representations. In this work, we consider the question: how can we leverage the compositional structure of \emph{language} for enabling agents to perform long-horizon tasks and systematically generalize to new goals?

To do so, we build upon the framework of hierarchical reinforcement learning (HRL), which offers a potential solution for learning long-horizon tasks by training a hierarchy of policies. However, the abstraction between these policies is critical for good performance.
Hard-coded abstractions often lack modeling flexibility and are task-specific~\citep{sutton1999between,konidaris2007building,heess2016learning, deepLoco}, while learned abstractions often find degenerate solutions without careful tuning~\citep{bacon2017option,harb2017waiting}.
One possible solution is to have the higher-level policy generate a sub-goal state and have the low-level policy try to reach that goal state~\citep{hiro,levy2018hierarchical}.
However, using goal states still lacks some degree of flexibility (e.g. in comparison to goal regions or attributes), is challenging to scale to visual observations naively, and does not generalize systematically to new goals.
In contrast to these prior approaches, language is a flexible representation for transferring a variety of ideas and intentions with minimal assumptions about the problem setting; its compositional nature makes it a powerful abstraction for representing combinatorial concepts and for transferring knowledge~\citep{grice1975logic}.

\begin{figure*}[tbp]
\centering
\begin{subfigure}[t]{0.32\textwidth}
\centering
\includegraphics[width=0.5\linewidth]{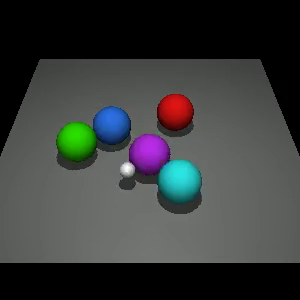}
\caption{\scriptsize Goal is $\vg_0$: ``{\it There is a \textbf{\textcolor{red}{red}} ball; are there any matte \textbf{\textcolor{cyan}{cyan}} sphere \textbf{right} of it?}". Currently $\Psi(\vs_t, \vg_0)=0$}
\end{subfigure}
~
\begin{subfigure}[t]{0.32\textwidth}
\centering
\includegraphics[width=0.5\linewidth]{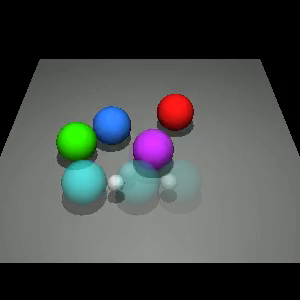}
\caption{\scriptsize Agent performs actions and interacts with the environment and tries to satisfy goal.
}
\end{subfigure}
~
\begin{subfigure}[t]{0.32\textwidth}
\centering
\includegraphics[width=0.5\linewidth]{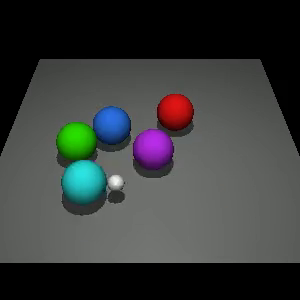}
\caption{\scriptsize Resulting state $\vs_{t+1}$ does not satisfy $\vg_0$,
so relabel goal to $\vg'$: ``{ \it There is a \textbf{\textcolor{green}{green}} sphere; are there any rubber \textbf{\textcolor{cyan}{cyan}} balls \textbf{behind} it?}" so $\Psi(\vs_{t+1}, \vg' )=1$}
\end{subfigure}
\vspace{-2mm}
\caption{\footnotesize The environment and some instructions that we consider in this work, along with an illustration of hindsight instruction relabeling (HIR), which we use to enable the agent to learn from many different language goals at once (Details in Section~\ref{sec:lowlevel}).}
\label{fig:hir_process}
\vspace{-3mm}
\end{figure*}

In this work, we propose to use language as the interface between high- and low-level policies in hierarchical RL.
With a low-level policy that follows language instructions (Figure \ref{fig:hir_process}), the high-level policy can produce actions in the space of language, yielding a number of appealing benefits.
First, the low-level policy can be re-used for different high-level objectives without retraining.
Second, the high-level policies are human-interpretable as the actions correspond to language instructions,
making it easier to recognize and diagnose failures.
Third, language abstractions can be viewed as a strict generalization of goal states, as an instruction can represent a \emph{region of states} that satisfy some abstract criteria, rather than the entirety of an individual goal state.
Finally, studies have also suggested that humans use language as an abstraction for reasoning and planning \citep{language_and_thought, piantadosi2012bootstrapping}. In fact, the majority of knowledge learning and skill acquisition we do throughout our life is through languages.

While language is an appealing choice as the abstraction for hierarchical RL, training a low-level policy to follow language instructions is highly non-trivial~\citep{fu2018from,bahdanau2018learning} as it involves learning from binary rewards that indicate completion of the instruction. To address this problem, we generalize prior work on goal relabeling to the space of language instructions (which instead operate on regions of state space, rather than a single state), allowing the agent to learn from many language instructions at once.

To empirically study the role of language abstractions for long-horizon tasks, we introduce a new environment inspired by the CLEVR engine~\citep{clevr} that consists of procedurally-generated scenes of objects that are paired with programatically-generated language descriptions. The low-level policy's objective is to manipulate the objects in the scene such that a description or statement is satisfied by the arrangement of objects in the scene.
We find that our approach is able to learn a variety of vision-based long-horizon manipulation tasks such as object reconfiguration and sorting, while outperforming state-of-the-art RL and hierarchical RL approaches. Further, our experimental analysis finds that HRL with non-compositional abstractions struggles to learn the tasks, even when the non-compositional abstraction is derived from language instructions themselves, demonstrating the critical role of compositionality in learning. Lastly, we find that our instruction-following agent is able to generalize to instructions that are systematically different from those seen during training.

In summary,
the main contribution of our work is three-fold:
\begin{enumerate}[leftmargin=1cm]
\vspace{-0.2cm}
     \item a framework for using language abstractions in HRL, with which we find that the structure and flexibility of language enables agents to solve challenging long-horizon control problems
     \item an open-source continuous control environment for studying compositional, long-horizon tasks, integrated with language instructions inspired by the CLEVR engine \citep{clevr}
     \item empirical analysis that studies the role of compositionality in learning long-horizon tasks and achieving systematic generalization
     \vspace{-0.2cm}
\end{enumerate}

\vspace{-0.1cm}
\section{Related Work}
\label{related}
\vspace{-0.1cm}

Designing, discovering and learning meaningful and effective abstractions of MDPs has been studied extensively in hierarchical reinforcement learning (HRL)~\citep{dayan1993feudal,parr1998reinforcement,sutton1999between,dietterich2000hierarchical,bacon2017option}.
Classically, the work on HRL has focused on learning only the high-level policy given a set of hand-engineered low-level policies~\citep{stolle2002learning,mannor2004dynamic,chentanez2005intrinsically}, or more generically \textit{options} policies with flexible termination conditions~\citep{sutton1999between,precup2000temporal}.

Recent HRL works have begun to tackle more difficult control domains with both large state spaces and long planning horizons~\citep{heess2016learning,kulkarni2016hierarchical,tessler2017deep,florensa2017stochastic,hiro,nachum2019near}. These works can typically be categorized into two approaches. The first aims to learn effective low-level policies end-to-end directly from final task rewards with minimal human engineering, such as through the option-critic architecture~\citep{bacon2017option,harb2017waiting} or multi-task or meta learning~\citep{frans2017meta,sigaud2018policy}. While appealing in theory, this end-to-end approach relies solely on final task rewards and is shown to scale poorly to complex domains~\citep{bacon2017option,hiro}, unless distributions of tasks are carefully designed~\citep{frans2017meta}.
The second approach instead augments the low-level learning with auxiliary rewards that can bring better inductive bias. These rewards include mutual information-based diversity rewards~\citep{daniel2012hierarchical,florensa2017stochastic}, hand-crafted rewards based on domain knowledge~\citep{konidaris2007building,heess2016learning,kulkarni2016hierarchical,tessler2017deep}, and goal-oriented rewards~\citep{dayan1993feudal,uvfa,her,vezhnevets2017feudal,hiro,nachum2019near}. Goal-oriented rewards have been shown to balance sufficient inductive bias for effective learning with minimal domain-specific engineering, and achieve performance gains on a range of domains~\citep{vezhnevets2017feudal,hiro,nachum2019near}. Our work is a generalization on these lines of work, representing goal \emph{regions} using language instructions, rather than individual goal \emph{states}. Here, \emph{region} refers to the sets of states (possibly disjoint and far away from each other) that satisfy more abstract criteria (e.g. \emph{``red ball in front of blue cube"} can be satisfied by infinitely many states that are drastically different from each other in the pixel space) rather than a simple $\epsilon$-ball around a single goal state that is only there for creating a reachable non-zero measure goal. Further, our experiments demonstrate significant empirical gains over these prior approaches.


Since our low-level policy training is related to goal-conditioned HRL, we can benefit from algorithmic advances in multi-goal reinforcement learning~\citep{kaelbling, uvfa, her, tdm}. In particular, we extend the recently popularized goal relabeling strategy~\citep{kaelbling, her} to instructions, allowing us to relabel based on achieving a language statement that describes a \emph{region} of state space, rather than relabeling based on reaching an individual state.


Lastly, there are a number of prior works that study how language can guide or improve reinforcement learning~\citep{luketina2019survey, fried2018speaker, kaplan2017beating, bahdanau2018learning, fu2018from,metalearning_language, babyai}. While prior work has made use of language-based sub-goal policies in hierarchical RL~\citep{shu2017hierarchical,das2018neural}, the instruction representations used lack the needed diversity to benefit from the compositionality of language over one-hot goal representations. 
In a concurrent work, \citet{advice} show that language can help with learning difficult tasks where more naive goal representations lead to poor performance, even with hindsight goal relabeling. While we are also interested in using language to improve learning of challenging tasks, we focus on the use of language in the context of hierarchical RL, demonstrating that language can be further used to compose complex objectives for the agent. 
\citet{andreasL3} leverage language descriptions to rapidly adapt to unseen environments through structured policy search in the space of language; each environment is described by one sentence. In contrast, we show that a high-level policy  can effectively leverage the combinatorially many sub-policies induced by language by generating a sequence of instructions for the low-level agent. Further, we use language not only for adaptation but also for learning the lower level control primitives, without the need for imitation learning from an expert. Another line of work focuses on RL for textual adventure games where the state is represented as language descriptions and the actions are either textual actions available at each state~\citep{he2015deep} or all possible actions~\citep{narasimhan2015language} (even though not every action is applicable to all states). In general, these works look at text-based games with discrete 1-bit actions, while we consider continuous actions in physics-based environments.
One may view the latter as a high-level policy with oracular low-level policies that are specific to each state; the discrete nature of these games entails limited complexity of interactions with the environment.
\vspace{-1mm}

\vspace{-0.1cm}
\section{Preliminaries}
\label{background}
\vspace{-0.1cm}
\label{sec:prelim}

\textbf{Standard reinforcement learning.}
The typical RL problem considers a {\it Markov decision process} (MDP) defined by the tuple  $(\gS, \gA, T, R, \gamma)$ where $\mathcal{S}$ where $\mathcal{S}$ is the state space, $\mathcal{A}$ is the action space,
the unknown transition probability $T: \mathcal{S} \times \mathcal{A} \times \mathcal{S} \rightarrow [0, \infty)$ represents the probability density of reaching $s_{t+1} \in \mathcal{S}$
from $\mathbf{s}_{t} \in \mathcal{S}$ by taking the action $\mathbf{a} \in \mathcal{A}$,
$\gamma \in [0, 1)$ is the discount factor,
and the bounded real-valued function $R: \mathcal{S} \times \mathcal{A} \rightarrow [r_{\min}, r_{\max}]$ represents the reward of each transition.
We further denote $\rho_\pi (\vs_t)$ and $\rho_\pi(\vs_t, \va_t)$ as the state marginal and the state-action marginal of the trajectory induced by policy $\pi(\va_t|\vs_t)$. 
The objective of  reinforcement learning is to find a policy $\pi(\va_t|\vs_t)$ such that the expected discounted future reward $\sum_t \E_{(\vs_t, \va_t)\sim \rho_\pi}[\gamma^t R(\vs_t, \va_t)]$ is maximized.

\textbf{Goal conditioned reinforcement learning.}
In goal-conditioned RL, we work with an
{\it Augmented Markov Decision Process},
which is defined by the tuple $(\mathcal{S}, \mathcal{G}, \mathcal{A}, T, R, \gamma)$. Most elements represent the same quantities as a standard MDP.
The additional tuple element $\mathcal{G}$ is the space of all possible goals, and the reward function $R: \mathcal{S} \times \mathcal{A} \times \mathcal{G} \rightarrow [r_{\min}, r_{\max}]$ represents the reward of each transition {\it under a given goal}.
Similarly, the policy $\pi(\va_t|\vs_t, \vg)$ is now conditioned on $\vg$.
Finally, $p_g(\vg)$ represents a distribution over $\gG$.
The objective of goal directed reinforcement learning is to find a policy $\pi(\va_t|\vs_t, \vg)$ such that the expected discounted future reward
$\sum_t \E_{\vg \sim p_g,(\vs_t, \va_t)\sim \rho_\pi}[\gamma^t R(\vs_t, \va_t, \vg)]$
is maximized. While this objective can be expressed with a standard MDP by augmenting the state vector with a goal vector, the policy does not change the goal; the explicit distinction between goal and state facilitates discussion later.

\textbf{Q-learning.}
Q-learning is a large class of off-policy reinforcement learning algorithms that focuses on learning the Q-function, $Q^*(\vs_t, \va_t)$,
which represents the expected total discounted reward that can be obtained after taking action $\va_t$ in state $\vs_t$ assuming the agent acts optimally thereafter. It can be recursively defined as:
\vspace{-0.15cm}
\begin{equation}
    Q^*(\vs_t, \va_t) = \E_{\vs_{t+1}} [R(\vs_t, \va_t) + \gamma \max_{\va \in \gA}(Q^*(\vs_{t+1}, \va))]
\end{equation}
The optimal policy learned can be recovered through $\pi^*(\va_t|\vs_t) = \delta(\va_t = \argmax_{\va\in\gA}Q^*(\vs_t, \va))$. 
In high-dimensional spaces, the Q-function is usually represented with function approximators and fit using transition tuples, $(\vs_t, \va_t, \vs_{t+1},
r_t)$, which are stored in a {\it replay buffer}~\cite{dqn}. 


\textbf{Hindsight experience replay (HER).}
HER \cite{her} is a data augmentation technique for off-policy goal conditioned reinforcement learning. For simplicity, assume that the goal is specified in the state space directly. A trajectory can be transformed into a sequence of goal augmented transition tuples $(\vs_t, \va_t, \vs_g, \vs_{t+1},
r_t)$. We can relabel each tuple's $\vs_g$ with $\vs_{t+1}$ or other future states visited in the trajectory and adjust $r_t$ to be the appropriate value.
This makes the otherwise sparse reward signal much denser. This technique can also been seen as generating an implicit curriculum of increasing difficulty for the agent as it learns to interact with the environment more effectively.
\vspace{-0.1cm}
\section{Hierarchical Reinforcement Learning with Language Abstractions}
\label{hrl_w_language}
\vspace{-0.1cm}

\begin{wrapfigure}{r}{0.33\textwidth}
\vspace{-5.9mm}
  \begin{center}
    \includegraphics[width=0.33\textwidth, clip]{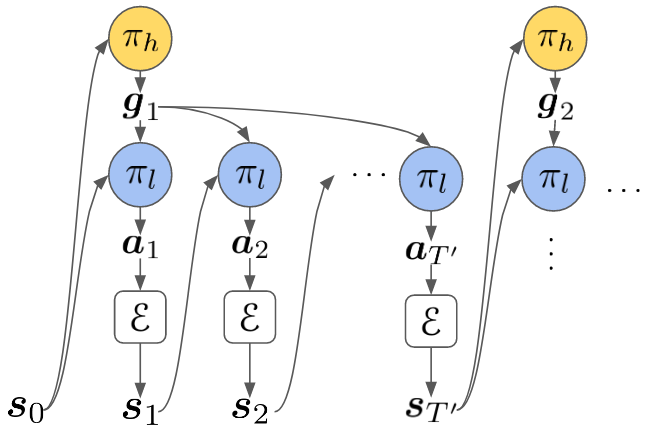}
  \end{center}
  \vspace{-0.4cm}
  \caption{\footnotesize HAL: The high-level policy $\pi_{h}$ produces language instructions $\vg$ for the low level policy $\pi_{l}$.}
  \vspace{-5mm}
\label{fig:overall_diagram}
\end{wrapfigure}

In this section, we present our framework
for training a 2-layer hierarchical policy with compositional \emph{language} as the abstraction between the high-level policy and the low-level policy. We open the exposition with formalizing the problem of solving temporally extended task with language, including our assumptions regarding the availability of supervision. We will then discuss how we can efficiently train the low-level policy, $\pi_{l}(\va | \vs_t, \vg)$ conditioned on language instructions $\vg$
in Section~\ref{sec:lowlevel}, and how a high-level policy, $\pi_{h}(\vg | \vs_t)$, can be trained using such a low-level policy in Section~\ref{sec:highlevel}. We refer to this framework as {\it Hierarchical Abstraction with Language} (HAL, Figure \ref{fig:overall_diagram}, Appendix \ref{app:overall_alg}).

\vspace{-0.2cm}
\subsection{Problem statement}
\vspace{-0.1cm}

We are interested in learning temporally-extended tasks by leveraging the compositionality of language. Thus, in addition to the standard reinforcement learning assumptions laid out in Section~\ref{sec:prelim}, we also need some form of grounded language supervision in the environemnt $\mathscr{E}$ during training. To this end, we also assume the access to a conditional density $\omega(\vg|\vs)$ that maps observation $\vs$ to a distribution of language statements $\vg \in \gG$ that describes $\vs$. This distribution can take the form of a supervised image captioning model, a human supervisor, or a functional program that is executed on $\vs_t$ similar to CLEVR. Further, we define $\Omega(\vs_t)$ to be the support of $\omega(\vg|\vs_t)$.
Moreover, we assume the access to a function $\Psi$ that maps a state and an instruction to a single Boolean value which indicates whether the instruction is satisfied or not by the $\vs$, i.e. $\Psi: \gS \times \gG \rightarrow \{0 ,1\}$. Once again, $\Psi$ can be a VQA model, a human supervisor or a program. Note that any goal specified in the state space can be easily expressed by a Boolean function of this form by checking if two states are close to each other up to some threshold parameter. $\Psi$ can effectively act as the reward for the low-level policy.

\begin{figure}[tbp]
\centering
\begin{subfigure}[t]{0.16\columnwidth}
\centering
\includegraphics[width=0.99\linewidth, height=0.8\linewidth]{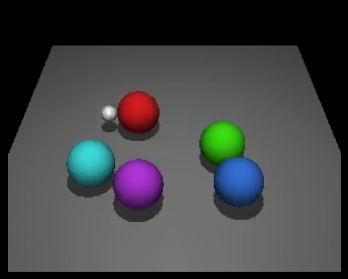}
\vspace{-0.5cm}
\caption{\footnotesize \centering Object \newline arrangement}
\end{subfigure}
\hfill
\begin{subfigure}[t]{0.16\columnwidth}
\centering
\includegraphics[width=0.99\linewidth, height=0.8\linewidth]{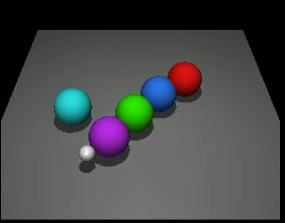}
\vspace{-0.5cm}
\caption{\footnotesize \centering Object \newline ordering}
\end{subfigure}
\hfill
\begin{subfigure}[t]{0.16\columnwidth}
\centering
\includegraphics[width=0.99\linewidth, height=0.8\linewidth]{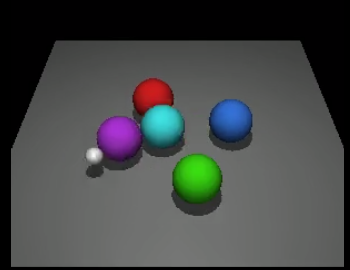}
\vspace{-0.5cm}
\caption{\footnotesize \centering Object \newline sorting}
\end{subfigure}
\hfill
\begin{subfigure}[t]{0.16\columnwidth}
\centering
\includegraphics[width=0.99\linewidth, height=0.8\linewidth]{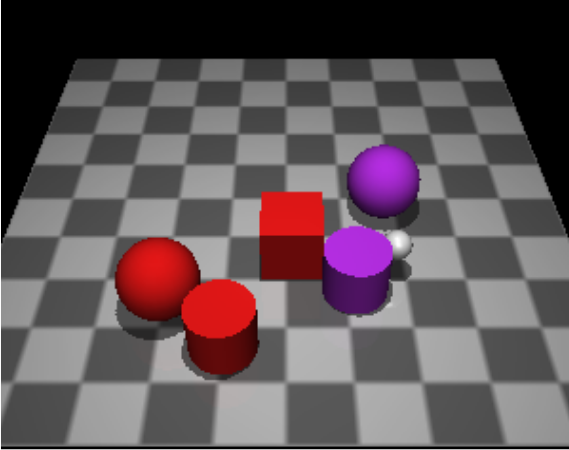}
\vspace{-0.5cm}
\caption{\footnotesize \centering Color\newline ordering}
\end{subfigure}
\hfill
\begin{subfigure}[t]{0.16\columnwidth}
\centering
\includegraphics[width=0.99\linewidth, height=0.8\linewidth]{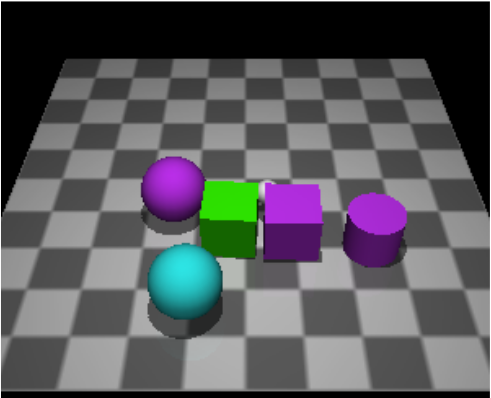}
\vspace{-0.5cm}
\caption{\footnotesize\centering Shape \newline ordering}
\end{subfigure}
\hfill
\begin{subfigure}[t]{0.16\columnwidth}
\centering
\includegraphics[width=0.99\linewidth, height=0.8\linewidth]{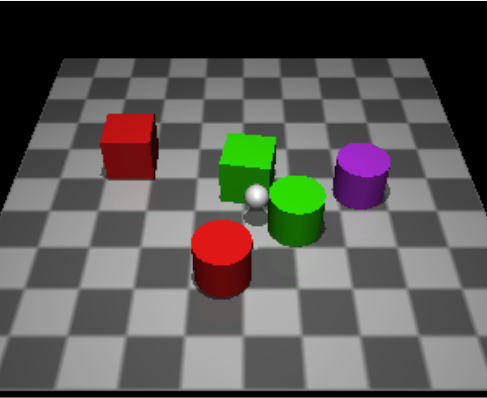}
\vspace{-0.5cm}
\caption{\footnotesize \centering Color \& shape ordering}
\end{subfigure}
\vspace{-0.1cm}
\caption{\footnotesize Sample goal states for the high-level tasks in the standard (a-c) and diverse (d-f) environments. The high-level policy only receives reward if all constraints are satisfied. The global location of the objects may vary.}
\vspace{-0.5cm}
\label{fig:highlevelgoal}
\end{figure}

An example for the high-level tasks is arranging objects in the scene according to a specific spatial relationship. This can be putting the object in specific arrangement or ordering the object according the the colors (Figure \ref{fig:highlevelgoal}) by pushing the objects around (Figure \ref{fig:hir_process}). Details of these high-level tasks are described in Section \ref{sec:env}. These tasks are complex but can be naturally decomposed into smaller sub-tasks, giving rise to a naturally defined hierarchy, making it an ideal testbed for HRL algorithms. Problems of similar nature including organizing a cluttered table top or stacking LEGO blocks to construct structures such as a castle or a bridge.
We train the low-level policy $\pi_l(\va|\vs, \vg)$ to solve an augmented MDP described Section~\ref{sec:prelim}.
For simplicity, we assume that $\Omega$'s output is uniform over $\gG$. The low-level policy receives supervision from $\Omega$ and $\Psi$ by completing instructions. The high-level policy $\pi_h(\vg|\vs)$ is trained to solve a standard MDP whose state space is the same $\gS$ as the low-level policy, and the action space is $\gG$. In this case, the high-level policy's supervision comes from the reward function of the environment which may be highly sparse.

We separately train the high-level policy and low-level policy, so the low-level policy is agnostic to the high-level policy. Since the policies share the same $\gG$, the low-level policy can be reused for different high-level policies (Appendix \ref{app:high_alg}). Jointly fine-tuning the low-level policy with a specific high-level policy is certainly a possible direction for future work (Appendix \ref{app:overall_alg}).


\vspace{-0.2cm}
\subsection{Training a language-conditioned low-level policy}
\label{sec:lowlevel}
\vspace{-0.1cm}



To train a goal conditioned low-level policy, we need to define a suitable reward function for training such a policy and a mechanism for sampling language instructions.
A straightforward way to represent the reward for the low-level policy would be $R(\vs_t, \va_t, \vs_{t+1}, \vg) = \Psi(\vs_{t+1}, \vg)$ or, to ensure that $\va_t$ is inducing the reward:
\vspace{-0.2cm}
\[
  R(\vs_t, \va_t, \vs_{t+1}, \vg) =
  \begin{cases}
                                   0 & \text{if $\Psi(\vs_{t+1}, \vg)=0$} \\
                                \Psi(\vs_{t+1}, \vg) 
                    \oplus\Psi(\vs_{t}, \vg) & \text{if $\Psi(\vs_{t+1}, \vg)=1$}
  \vspace{-0.0cm}
  \end{cases}
\]
However, optimizing with this reward directly is difficult because the reward signal is only non-zero when the goal is achieved. Unlike prior work (e.g. HER \cite{her}),
which uses a state vector or a task-relevant part of the state vector as the goal, it is difficult to define meaningful distance metrics in the space of language statements \citep{callison2006re, reiter2018structured, sulem2018bleu}, and, consequently, difficult to make the reward signal smooth by assigning partial credit (unlike, e.g., the $\ell_p$ norm of the difference between 2 states). 
To overcome these difficulties, we propose a trajectory relabeling technique for language instructions:
Instead of relabeling the trajectory with states reached,
we relabel states in the the trajectory $\tau$ with the elements of $\Omega(\vs_t)$ as the goal instruction using a relabeling strategy $\mathscr{S}$. 
We refer to this procedure as {\it hindsight instruction relabeling} (HIR). The details of $\mathscr{S}$ is located in Algorithm~\ref{app:relabel_alg} in Appendix~\ref{app:relabel}. 
Pseudocode for the method can be found in Algorithm~\ref{alg:hir} in Appendix~\ref{app:low_alg} and an illustration of the process can be found in Figure \ref{fig:hir_process}.

The proposed relabeling scheme, HIR, 
is reminiscent of HER \citep{her}.
In HER, the goal is often the state or a simple function of the state, such as a masked representation.
However, with high-dimensional observation spaces such as images, there is excessive information in the state that is irrelevant to the goal, while task-relevant information is not readily accessible. While one can use HER with generative models of images \citep{visual_her,skewfit,imagined}, the resulting representation of the state may not effectively capture the relevant aspects of the desired task. Language can be viewed as an alternative, highly-compressed representation of the state that explicitly exposes the task structure, e.g. the relation between objects. Thus, we can readily apply HIR to settings with image observations.

\vspace{-0.2cm}
\subsection{Acting in language with the high-level policy}
\vspace{-0.1cm}
\label{sec:highlevel}
We aim to learn a high-level policy for long-horizon tasks that can explore and act in the space of language by providing instructions $\vg$ to the low-level policy $\pi_{l}(\va | \vs_t, \vg)$. The use of language abstractions through $\vg$ allows the high-level policy to structurally explore with actions that are semantically meaningful and span multiple low-level actions. 

In principle, the high-level policy, $\pi_{h}(\vg|\vs)$, can be trained with any reinforcement learning algorithm,
given a suitable way to generate sentences for the goals.
However, generating coherent sequences of discrete tokens is difficult, \change{particularly when combined with existing reinforcement learning algorithms. We explore how we might incorporate a language model into the high level policy in Appendix~\ref{sec:LM}, which shows promising preliminary results but also significant challenges.}
Fortunately, while the size of the instruction space $\gG$ scales exponentially with the size of the vocabulary, the elements of $\gG$ are naturally structured and redundant -- many elements correspond to effectively the same underlying instruction with different synonyms or grammar. While the low-level policy understands all the different instructions, in many cases, the high-level policy only needs to generate instruction from a much smaller subset of $\gG$ to direct the low-level policy. We denote such subsets of $\gG$ as $\gI$.

If  $\gI$ is relatively small,
the problem can be recast as a discrete-action RL problem, where one action choice corresponds to an instruction,
and can be solved with algorithms such as DQN \cite{dqn}. We adopt this simple approach in this work.
As the instruction often represents a sequence of low-level actions, we take $T'$ actions with the low-level policy for every high-level instruction. $T'$ can be a fixed number of steps, or computed dynamically by a terminal policy learned by the low-level policy like the option framework. We found that simply using a fixed $T'$ was sufficient in our experiments.

\vspace{-0.2cm}
\section{The Environment and Implementation}
\label{sec:env}
\vspace{-0.1cm}

\textbf{Environment.} 
To empirically study how compositional languages can aid in long-horizon reasoning and generalization, we need an environment that will test the agent's ability to do so.
While prior works have studied the use of language in navigation~\citep{vln}, instruction following in a grid-world~\citep{babyai}, and compositionality in question-answering, we aim to develop a \emph{physical} simulation environment where the agent must interact with and change the environment in order to accomplish long-horizon, compositional tasks.
These criteria are particularly appealing for robotic applications, and, to the best of our knowledge, none of the existing benchmarks simultaneously fulfills all of them.
To this end, we developed a new environment using the MuJoCo physics engine~\citep{todorov2012mujoco} and the CLEVR language engine, that tests an agents ability to manipulate and rearrange objects of various shapes and colors.
To succeed in this environment, the agent must be able to handle varying number of objects with diverse visual and physical properties. 
Two versions of the environment of varying complexity are illustrated in Figures~\ref{fig:highlevelgoal} and~\ref{fig:hir_process} and further details are in Appendix~\ref{app:env}.

\textbf{High-level tasks.}
We evaluate our framework 6 challenging temporally-extended tasks across two environments, 
all illustrated in Figure \ref{fig:highlevelgoal}: (a) \emph{object arrangement}: manipulate objects such that 10 pair-wise constraints are satisfied, (b) \emph{object ordering}: order objects by color, (c) \emph{object sorting}: arrange 4 objects around a central object, and in a more diverse environment (d) \emph{color ordering}: order objects by color irrespective of shape, (e) \emph{shape ordering}: order objects by shape irrespective of color, and (f) \emph{color \& shape ordering}: order objects by both shape and color.
In all cases, the agent receives a binary reward only if all constraints are satisfied. Consequently, this makes obtaining meaningful signal in these tasks extremely challenging as only a very small number of action sequences will yield non-zero signal. For more details, see Appendix~\ref{app:task}.

\textbf{Action and observation parameterization.} The state-based observation is $\vs \in \sR^{10}$ that represents the location of each object and $|\gA| = 40$ which corresponds to picking an object and pushing it in one of the eight cardinal directions. The image-based observation is $\vs \in \sR^{64 \times 64 \times 3}$ which is the rendering of the scene and $|\gA| = 800$ which corresponds to picking a location in a $10\times10$ grid and pushing in one of the eight cardinal directions. For more details, see Appendix~\ref{app:env}.

\textbf{Policy parameterization.} The low-level policy encodes the instruction with a GRU and feeds the result, along with the state, into a neural network that predicts the Q-value of each action.
The high-level policy is also a neural network Q-function. Both use Double DQN \cite{doubleq} for training.
The high-level policy uses sets of 80 and 240 instructions as the action space in the standard and diverse environments respectively, a set that sufficiently covers relationships between objects. 
We roll out the low-level policy for $T'=5$ steps for every high-level instruction. For details, see Appendix~\ref{app:impl}.
\vspace{-0.2cm}
\section{Experiments}
\vspace{-0.1cm}



To evaluate our framework, and study the role of compositionality in RL in general, we design the experiments to answer the following questions: 
\textbf{(1)} as a representation, how does language compare to alternative representations, such as those that are not explicitly compositional?
\textbf{(2)} How well does the framework scale with the diversity of instruction and dimensionality of the state (e.g. vision-based observation)?
\textbf{(3)} Can the policy generalize in systematic ways by leveraging the structure of language?
\textbf{(4)} Overall, how does our approach compare to state-of-the-art hierarchical RL approaches, along with learning flat, homogeneous policies?

\change{To answer these questions,} in Section~\ref{sec:lowlevel-exp}, we \change{first} evaluate and analyze training of effective low-level policies, which are critical for effective learning of long-horizon tasks. \change{Then}, in Section~\ref{sec:high-level-exp}, we evaluate the full HAL method on challenging temporally-extended tasks. \change{Finally, we apply our method to the Crafting environment from \citet{psketch} to showcase the generality of our framework.} For details on the experimental set-up and analysis, see Appendix~\ref{app:exp_detail} and \ref{app:analysis}.

\vspace{-0.1cm}
\subsection{Low-level Policy}\label{sec:lowlevel-exp}
\vspace{-0.1cm}

\begin{figure*}[tbp]
\centering
\begin{subfigure}[t]{0.32\textwidth}
\centering
\includegraphics[height=0.75\textwidth, clip]{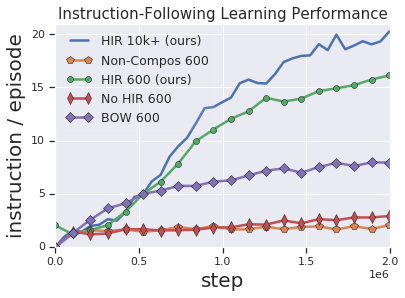}
\end{subfigure}
~
\begin{subfigure}[t]{0.32\textwidth}
\centering
\includegraphics[ height=0.75\textwidth, clip]{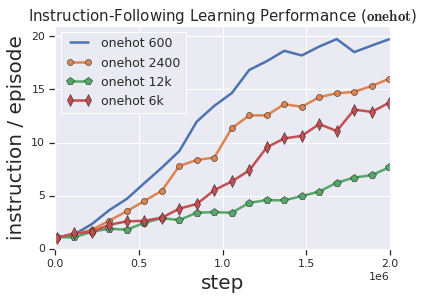}
\end{subfigure}
~
\begin{subfigure}[t]{0.32\textwidth}
\centering
\includegraphics[height=0.75\textwidth, clip]{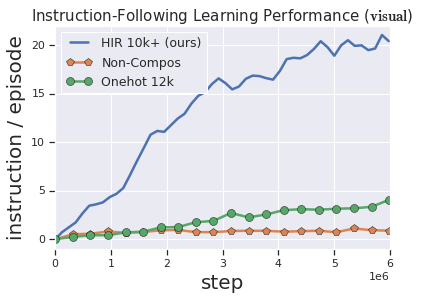}
\end{subfigure}

\vspace{-0.3cm}
\caption{
\footnotesize 
Results for low-level policies in terms of goals accomplished per episode over training steps for HIR. \textbf{Left}: HIR with different number of instructions and results with non-compositional representation and with no relabeling. \textbf{Middle}: Results for one-hot encoded representation with increasing number of instructions. Since the one-hot cannot leverage compositionality of the language, it suffers significantly as instruction sets grow, while HIR on sentences in fact learns even faster when instruction sets increase. \textbf{Right}: Performance of image-based low-level policy compared against one-hot and non-compositional instruction representations.}
\vspace{-0.3cm}
\label{fig:lowlevel}
\end{figure*}

\textbf{Role of compositionality and relabeling.}
We start by evaluating the fidelity of the low-level instruction-following policy, in isolation, with a variety of representations for the instruction. For these experiments, we use state-based observations. We start with a set of 600 instructions, which we  paraphrase and substitute synonyms to obtain more than 10,000 total instructions which allows us to answer the first part of \textbf{(2)}. We evaluate the performance all low-level policies by the average number of instructions it can successfully achieve each episode (100 steps), measured over 100 episodes. To answer \textbf{(1)}, and evaluate the importance of compositionality, we compare against:
\vspace{-0.1cm}
\begin{itemize}[leftmargin=*]
    \item a \textbf{one-hot encoded representation} of instructions where each instruction has its own row in a real-valued embedding matrix which uses the same instruction relabeling (see Appendix \ref{app:ohe})
    \vspace{-0.05cm}
    \item a \textbf{non-compositional latent variable representation} with identical information content. We train an sequence auto-encoder on sentences, which achieves 0 reconstruction error and is hence a
    \vspace{-0.05cm}
    \emph{lossless} non-compositional representation of the instructions (see Appendix \ref{app:nc})
    \item \change{a \textbf{bag-of-words (BOW) representation} of instructions (see Appendix \ref{app:bow})}
\end{itemize}
\vspace{-0.1cm}
In the first comparison, we observe that while one-hot encoded representation works on-par with or better than the language in the regime where the number of instructions is small, its performance quickly deteriorates as the number of instruction increases (Fig.\ref{fig:lowlevel}, middle). On the other hand, language representation of instruction can leverage the structures shared by different instructions and does not suffer from increasing number of instructions (Fig.\ref{fig:lowlevel}, right blue); in fact, an \emph{improvement} in performance is observed. This suggests, perhaps unsurprisingly, that one-hot representations and state-based relabeling scale poorly to large numbers of instructions, \emph{even when the underlying number of instructions does not change}, while, with instruction relabeling (HIR), the policy acquires better, more successful representations as the number of instructions increases.

In the second comparison, we observe that the agent is unable to make meaningful progress with this representation despite receiving identical amount of supervision as language. This indicates that the compositionality of language is critical for effective learning.
Finally, we find that relabeling is critical for good performance, since without it (no HIR), the reward signal is significantly sparser.

\change{Finally, in the comparison to a bag-of-words representation (BOW), we observe that, while the BOW agent's return increases at a faster rate than that of the language agent at the beginning of training -- likely due to the difficulty of optimizing recurrent neural network in language agent -- the language agent achieves significantly better final performance. On the other hand, the performance of the BOW agent plateaus at around 8 instructions per episode. This is expected as BOW does not consider the {\it sequential nature} of an instruction, which is important for correctly executing an instruction.}

\textbf{Vision-based observations.}
To answer the second part of \textbf{(2)}, we extend our framework to pixel observations. The agent reaches the same performance as the state-based model, albeit requiring longer convergence time with the same hyper-parameters. On the other hand, the one-hot representation reaches much worse relative performance with the same amount of experience (Fig.\ref{fig:lowlevel}, right).


\textbf{Visual generalization.}
One of the most appealing aspects of language is the promise of {\it combinatorial generalization} \cite{battaglia2018relational} which allows for \emph{extrapolation} rather than simple \emph{interpolation} over the training set. 
To evaluate this (i.e. \textbf{(3)}), we design the training and test instruction sets that are systematically distinct. 
We evaluate the agent's ability to perform such generalization by splitting the 600 instruction sets through the following procedure: \textbf{(i)} \textbf{standard}: random 70/30 split of the instruction set; \textbf{(ii)} \textbf{systematic}: the training set only consists of instructions that do not contain the words {\it red} in the first half of the instructions and the test set contains only those that have  {\it red} in the first half of the instructions. We emphasize that the agent has {\it never} seen the words red in the first part of the sentence in training; in other words, the task is \emph{zero-shot} as the training set and the test set are disjoint (i.e. the distributions do not share support). From a pure statistical learning theoretical perspective, the agent should not do better than chance on such a test set. Remarkably, we observe that the agent generalizes better with language than with non-compositional representation (table \ref{tab:generalization}). This suggests that the agent recognizes the compositional structure of the language, and achieves systematic generalization through such understanding.

\begin{table}[htbp]
\centering
\begin{tabular}{p{1.9cm}cccccc} \toprule
 & {\scriptsize \textbf{\makecell{Standard \\ train}}} & {\scriptsize \textbf{\makecell{Standard \\ test}}} & {\scriptsize \textbf{\makecell{Standard \\ gap}}} & {\scriptsize \textbf{\makecell{Systematic \\ train}}} & {\scriptsize \textbf{\makecell{Systematic\\test}}} & {\scriptsize \textbf{\makecell{Systematic \\ gap}}}\\ \midrule
\scriptsize \textbf{\!\!\!Language} & \scriptsize{$\boldsymbol{21.50 \pm 2.28}$} & \scriptsize{$\boldsymbol{21.49 \pm 2.53}$} & \scriptsize{$\boldsymbol{0.001}$} & \scriptsize{$\boldsymbol{20.09 \pm 2.46}$} & \scriptsize{$\boldsymbol{8.13 \pm 2.34}$\!\!} & \scriptsize{$\boldsymbol{0.596}$}\\ \midrule
\scriptsize{\textbf{\!\!\!\mbox{Non-Compositional}}\!\!} & \scriptsize{\scriptsize{$6.26 \pm 1.18$}} & \scriptsize{\scriptsize{$5.78 \pm 1.44$}} & \scriptsize{$0.077$} & \scriptsize{$7.54 \pm 1.14$} & \scriptsize{$0.76 \pm 0.69$\!\!} & \scriptsize{$0.899$}\\ \midrule
\scriptsize \textbf{\!\!\!Random} & \scriptsize{ $0.17 \pm 0.20$} & \scriptsize{ $0.21 \pm 0.17$} & \scriptsize{-} & \scriptsize{ $0.11 \pm 0.19$} & \scriptsize{ $0.18 \pm 0.22$\!\!} & \scriptsize{-} \\ \bottomrule
\end{tabular}
\vspace{3mm}
\caption{\footnotesize Final performance of the low-level policy on different training and test instruction distributions (20 episodes). Language outperforms the non-compositional language representation in both absolute performance and relative generalization gap for every setting. \emph{Gap} is equal to one minus the ratio of between mean test performance and mean train performance; this quantity can be interpreted as the generalization gap. For instructions with language representation, the generalization gap increases by approximately $59.5\%$ from standard generalization to zero-shot generalization while for non-compositional representation the generalization gap increases by $82.2\%$}
\label{tab:generalization}
\vspace{-5mm}
\end{table}

\begin{figure}[htbp]
\centering
\begin{subfigure}[t]{0.31\textwidth}
\centering
\includegraphics[height=0.75\textwidth]{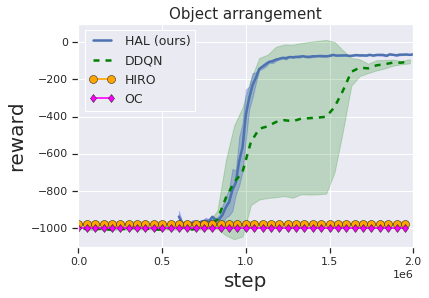}
\end{subfigure}
~
\begin{subfigure}[b]{0.31\textwidth}
\centering
\includegraphics[height=0.75\textwidth]{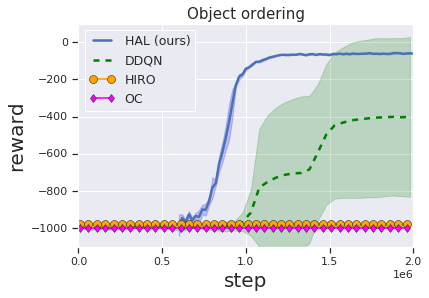}
\end{subfigure}
~
\begin{subfigure}[t]{0.31\textwidth}
\centering
\includegraphics[height=0.75\textwidth]{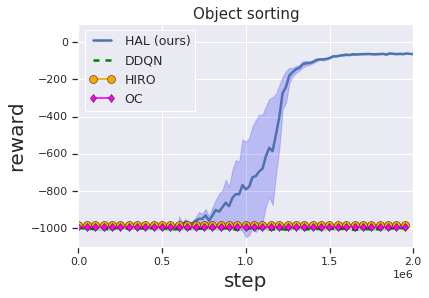}
\end{subfigure}
\vspace{-0.1cm}
\caption{\footnotesize Results for high-level policy on tasks (a-c). Blue curves for HAL include the steps for training the low-level policy (a \emph{single} low-level policy is used for all 3 tasks).
In all settings, HAL demonstrates faster learning than DDQN. Means and standard deviations of 3 random seeds are plotted.}
\label{fig:highlevel}
\vspace{-0.2cm}
\end{figure}

\vspace{-0.2cm}
\subsection{High-level policy}\label{sec:high-level-exp}
\begin{figure}[htbp]
\centering
\begin{subfigure}[t]{0.31\textwidth}
\centering
\includegraphics[height=0.85\textwidth]{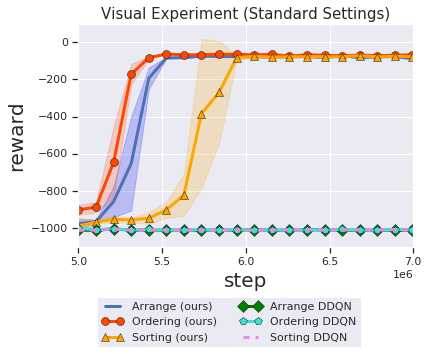}
\end{subfigure}
\hspace{-0.2cm}
\begin{subfigure}[b]{0.31\textwidth}
\centering
\includegraphics[height=0.85\textwidth]{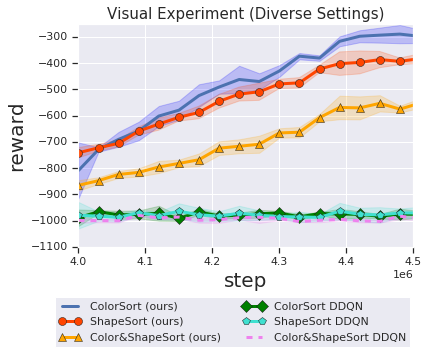}
\end{subfigure}
\hspace{-0.2cm}
\begin{subfigure}[t]{0.31\textwidth}
\centering
\includegraphics[height=0.82\textwidth]{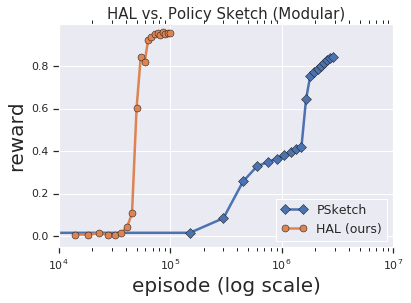}
\end{subfigure}
\vspace{-0.1cm}
\caption{\footnotesize \textbf{Left}: Results for \textbf{vision-based} hierarchical RL. In all settings, HAL demonstrates faster and more stable learning while the baseline DDQN cannot learn a non-trivial policy. In this case, the vision-based low-level policy needs longer training time (\textasciitilde$5\times10^6$ steps) so we start the x-axis there. Means and standard deviations of 3 seeds are plotted (Near 0 variance for DDQN). \textbf{Middle}: Results of HRL on the proposed 3 diverse tasks (d-e). In this case, the low-level policy used is trained on image observation for (\textasciitilde$4\times10^6$ steps). 3 random seeds are plotted and training has not converged. \change{\textbf{Right}: HAL vs policy sketches on the Crafting Environment. HAL is significantly more sample efficient since it is off-policy and uses relabeling among all modules.}}
\label{fig:vision_highlevel}
\vspace{-0.3cm}
\end{figure}

Now that we have analyzed the low-level policy performance, we next evaluate the full HAL algorithm.
To answer \textbf{(4)}, we compare our framework in the state space against a non-hierarchical baseline DDQN and two representative hierarchical reinforcement learning frameworks HIRO~\cite{hiro} and Option-Critic (OC) \cite{bacon2017option} on the proposed high-level tasks with sparse rewards (Sec.\ref{sec:env}). 
We observe that neither HRL baselines are able to learn a reasonable policy while DDQN is able to solve only 2 of the 3 tasks. HAL is able to solve all 3 tasks consistently and with much lower variance and better asymptotic performance (Fig.\ref{fig:highlevel}). Then we show that our framework successfully transfers to high-dimensional observation (i.e. images) in all 3 tasks without loss of performance whereas even the non-hierarchical DDQN fails to make progress (Fig.\ref{fig:vision_highlevel}, left). Finally, we apply the method to 3 additional diverse tasks (Fig.\ref{fig:vision_highlevel}, middle). In these settings, we observed that the high-level policy has difficulty learning from pixels alone, likely due to the visual diversity and the simplified high-level policy parameterization. As such, the high-level policy for diverse setting receives state observation but the low-level policy uses the raw-pixel observation. For more details, please refer to Appendix \ref{app:impl}.


\vspace{-0.1cm}
\subsection{Crafting Environment}\label{sec:crafting-exp}
\vspace{-0.1cm}

\change{To show the generality of the proposed framework, we apply our method to the \textbf{Crafting} environments introduced by \citet{psketch} (Fig.~\ref{fig:vision_highlevel}, right). We apply HAL to this environment by training {\it separate} policy networks for each module since there are fewer than 20 modules in the environment. These low-level policies receive binary rewards (i.e. one-bit supervision analogous to completing an instruction), and are trained jointly with HIR. Another high-level policy that picks which module to execute for a fixed 5 steps and is trained with regular DDQN. Note that our method uses a \textit{different} form of supervision compared to policy sketch since we provide binary reward for the low-level policy -- such supervision can sometimes be easier to specify than the entire sketch.}

\section{Discussion}\label{discussion}
\vspace{-0.1cm}

We demonstrate that language abstractions can serve as an efficient, flexible,
and human-interpretable representation for solving a variety of long-horizon control problems in HRL framework. Through relabeling and the inherent compositionality of language, we show that low-level, language-conditioned policies can be trained efficiently without engineered reward shaping and with large numbers of instructions, while exhibiting strong generalizations.
Our framework HAL can thereby leverage these policies to solve a range of difficult sparse-reward manipulation tasks with greater success and sample efficiency than training without language abstractions.

While our method demonstrates promising results, one limitation is that the current method relies on instructions provided by a language supervisor which has access to the instructions that describe a scene. The language supervisor can, in principle, be replaced with an image-captioning model and question-answering model such that it can be deployed on real image observations for robotic control tasks, an exciting direction for future work.
Another limitation is that the instruction set used is specific to our problem domain, providing a substantial amount of pre-defined structure to the agent. It remains an open question on how to enable an agent to follow a much more diverse instruction set, that is not specific to any particular domain, or learn compositional abstractions without the supervision of language. Our experiments suggest that both would likely yield an HRL method that requires minimal domain-specific supervision, while yielding significant empirical gains over existing domain-agnostic works, indicating an promising direction for future research.
Overall, we believe this work represents a step towards RL agents that can effectively reason using compositional language to perform complex tasks, and hope that our empirical analysis will inspire more research in compositionality at the intersection of language and reinforcement learning.

\subsubsection*{Acknowledgments}
We would like to thank anonymous reviewers for the valuable feedback. We would also like to thank Jacob Andreas, Justin Fu, Sergio Guadarrama, Ofir Nachum, Vikash Kumar, Allan Zhou, Archit Sharma, and other colleagues at Google Research for helpful discussion and feedback on the draft of this work.

\bibliography{main}
\bibliographystyle{plainnat}

\clearpage
\appendix
\section{Language Model as the High-level Policy} \label{sec:LM}
Picking an instruction from a pre-defined set of instructions can limit the flexibility of the high-level policy and introduces scaling issues for more complex language instructions. In this section, we present a method for using a language model as the high-level policy, thereby removing the constraints of fixed instruction set and potentially enabling much more powerful and general-purpose high-level policy. Naively, directly using language as the action space is difficult for RL agent as language has extremely high-dimension and presents difficult exploration challenges. An insight that we can leverage is that many language statements, especially instructions, lie in a much lower dimension space. To that end, we can learn a low-dimension embedding space of the instructions and then set the action space of the high-level RL agent to be the embedding space. Note that naively training a latent variable model such as a VAE~\citep{kingma2013auto} is unlikely work well since the latent space does not have the structure necessary for being a good action space -- an indirect example that demonstrates this fact is that the non-compositional representation for the low-level policy, which is also a latent variable model completely fails to learn a reasonable policy, even though it contains perfect information about the instruction. We hypothesize that to facilitate learning, the embedding space must be sufficiently \textit{disentangled}. Learning a disentangled representation of language without any supervision is very much still an open problem and we present a candidate solution that can learn an well disentangled embedding space (for a relatively domain-specific language) that can efficiently be used as the action space for the high-level policy.

\subsection{Learning a Disentangled Representation of Language}
Unlike modeling the density of language with architectures such as GPT or Transformers, we want to learn an disentangled representation of language via an embedding space. To this end, we leverage an InfoGAN-like auto-regressive architecture that turns an embedding to a sentence, and use a mutual information based objective to encourage disentanglement. Further, since the generated tokens are discrete, we explore several method for efficiently training the generator with gradients and find that using vector quantization~\citep{vqvae} empirically outperforms unbiased methods such as the Gumbel-softmax trick~\citep{jang2016categorical}.

\textbf{Notations}:
\begin{enumerate}
    \item The latent vector $\widetilde{\vz} = (\vz, \vc)\in \sR^{10}$ is a concatenation of a noise vector $\vz \in \sR^4$ sampled from an isotropic Gaussian and a latent code $\vc \in \sR^6$ sampled uniformly from the hypercube $[-1, 1]^6$. With a little abuse of notations, we use $\widetilde{\vz}_{\vc}$ to denote the latent code components of the latent vector $\widetilde{\vz}$.
    \item $\gV = \{v_i\}_{i=1}^{N_{\gV}}$ is the set of all vocabularies with size $N_{\gV}$
    \item $\gE = \{\ve_i\}_{i=1}^{N_{\gV}}$ is a set embedding vector for each vocabulary where $\ve \in \sR^{d_{e}}$. Further, for stability, each $\ve_i$ is normalize to be unit vector such that $||\ve_i||_2=1$ (constrained to a unit sphere).
    \item $\vx = [v_1, v_2, \dots, v_{m}]$ is real instruction datum, assuming all instructions are of length $m$ (padded with $\langle EOS \rangle$ if shorter).
    \item $\vx_i$ is the $i^{th}$ example in a collection of sampled instructions $\gX = \{\vx_i\}_{i=1}^{N_{\gX}}$.
    \item $G(\widetilde{\vz}; \theta_G)$ is the generator that maps a latent code to a vector $\vx_{gen} \in \sR^{m \times d_{e}}$. Each row of $\vx_{gen}^{(i)}$ can be mapped to some
    $$\ve(\vx_{gen}^{(i)})=\argmin_{\ve_j}||\ve_j-\vx_{gen}^{(i)}||_2$$
    \item $D(\vx\,;\theta_D)$ maps a sample to a scalar; $D$ is constrained to be 1-Lipschitz.
    \item $Q(\vx_{gen}\,;\theta_Q)$ maps a generated sample to its latent code $\widetilde{\vc}$. $Q$ shares the parameters all layers except for the last layer with $D$
    \item sg denotes stop gradient operator
\end{enumerate}

Then the \textbf{training objective of the generator} is defined as:
\begin{equation}
\ell_{VQ}(\widetilde{\vz}) = \frac{1}{m}\sum_{i=1}^{m}||\;\textrm{sg}[G(\widetilde{\vz})^{(i)}] - \ve(G(\widetilde{\vz})^{(i)})\,||_2^2 + \beta ||\,G(\widetilde{\vz})^{(i)} - \textrm{sg}[\ve(G(\widetilde{\vz})^{(i)})]\,||_2^2 
\end{equation}
\begin{equation}
\ell_G = \E_{\widetilde{\vz}} [ -D(G(\widetilde{\vz})) + \textrm{huber}(Q(G(\widetilde{\vz})), \, \widetilde{\vz}_{\vc}) + \ell_{VQ}(\widetilde{\vz})]
\end{equation}
\begin{equation}
(\gE^*, \theta_{G}^*) = \argmin_{\gE, \theta_{G}} \ell_G
\end{equation}
$\beta$ is a hyperparameter for the \emph{commitment loss} (second term in the VQ objective) and huber denotes the Huber loss. Note that Huber loss is used instead of $\ell_2$ loss which corresponds to true maximum likelihood objective as $\ell_2$ loss is very unstable.

Then the \textbf{training objective of the discriminator} is defined as:
\begin{equation}
\ell_D = \E_{\widetilde{\vz}}[\, \max(0, 1 + D(G(\widetilde{\vz}))) \,] + \E_{\vx \sim\gU(\gX)} [\,\max(0, 1 - D(\vx))\,]
\end{equation}
\begin{equation}
\theta_{D}^* = \argmin_{\theta_{D} \,\, \textrm{s.t} \,\, D \in C^{1,1}} \ell_D
\end{equation}
Note that the discriminator does not optimize $\gE$. Including $\gE$ causes instability likely from competition between the generator and discriminator. The generator \textit{cannot} cheat by forcing all embeddings to be the same (which will fool the discriminator that does not have control over the embedding) because of the mutual information constraint below.

The \textbf{training objective of the decoder} is defined as:
\begin{equation}
\ell_Q = \E_{\widetilde{\vz}}[\, \textrm{huber}(Q(G(\widetilde{\vz})), \, \widetilde{\vz}_\vc) \, ]
\end{equation}
\begin{equation}
\theta_{Q}^* = \argmin_{\theta_{Q}} \ell_Q
\end{equation}
For the components shared with D, the 1-Lipschitz constraint is also enforced.

\subsection{Embedding as Action}
After training the architecture above, we have a function G that maps a noise vector $\widetilde{\vz}$ to a sentence. We train a SAC \citep{haarnoja2018soft} agent $\pi_h$ that outputs action $\va \in \sR^{10}$. This action is mapped to a instruction $\vg = G(\va)$ which is in turn interpreted and executed by the low-level policy $\pi_l$. The policy is trained as a regular reinforcement learning agent on the high-level tasks.

\subsection{Architecture Details}
The generator network consists of a GRU cell with 128 hidden units. There are 50 available tokens for all the sentences in the environment and we choose the embedding size to be 64 so the embedding vector is of dimension $50\times32$. The hidden state at each time step of the generator is linearly projected to dimension 32.

The discriminator architecture is a feed-forward 1D convolutional neural network with residual connections.
The components shared by the discriminator and the decoder consists of 2 identical residual blocks, each of which consists of an identity block plus the output of convolution with window size 1 and output channel 32, leaky relu, convolution with window size 3 and output channel 64, and leaky relu. All of the convolution layers are spectrally normalized \citep{miyato2018spectral}. Following these are branches of two identical residual blocks -- one for the discriminator and the other for the decoder. Finally, the outputs of both are reduced to a single scalar.

\subsection{Experiments}
We trained the above language model with 10000 randomly sampled instructions and trained a SAC agent using the resulting generator on the ColorSort task from the diverse setting. The diverse setting has much more possible instructions compared to the fixed object setting and can benefit more from having access to more instructions at the high-level policy. The comparison between language model and fixed instruction set (Discrete) is shown below in Figure \ref{fig:lm}.

\begin{figure}[!htbp]
\centering
\centering
\includegraphics[width=0.5\textwidth]{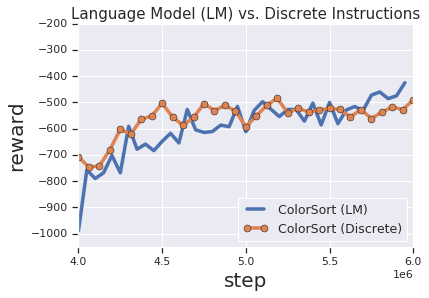}

\caption{\footnotesize Results of an agent using a language model vs an agent picking from a fixed set of instructions.}
\label{fig:lm}
\vspace{-0.3cm}
\end{figure}

Currently the language model agent performs comparably to the discrete agent. This may be due the fact that learning a disentangled generative model of language is difficult -- in particular, we observed that the model tends to drop modes which can affect the final policy negatively. We expect the result to significantly improve with better language models or even visual-linguistic models. Further, the fixed instruction baseline shown above under-performs the one used in the main text because it \textit{does not} use prioritized replay but instead uses uniform replay buffer. We observed decreased performance for SAC when prioritized replay is used; as such, we decided to use a uniform replay buffer for fair comparison.

\newpage

\section{Environment, Model Architectures, and Implementation}

In this section, we describe the environment and tasks we designed, and various implementation details such as the architectures of the policy networks. 

\subsection{CLEVR-Robot Environment}
\label{app:env}

We want an environment where we can evaluate the agent's ability to learn policies for long-horizon tasks that have compositional structure.
In robotics, manipulating and rearranging objects is a fundamental way through which robots interact with the environment, which is often cluttered and unstructured.
Humans also tend to decompose complex tasks into smaller sub-tasks in the environment (e.g. putting together a complicated model or write a piece of program).
While previous works have studied the use of language in navigation domains, we aim to develop an environment where the agent can physically interact with and change the environment. To that end, we designed a new environment for language and manipulation tasks in MuJoCo where the agents must interact with the objects in the scene.
To succeed in this environment, the agents must be able to handle a variety of objects with diverse visual and physical properties.
Since our environment is inspired by the CLEVR dataset, we name our environment \emph{CLEVR-Robot Environment}.

We will refer to all the elements in the environment collectively as the {\it world state}. The environment can contain up to 5 object. Each object is represented by a vector $o_i$ that is the concatenation of {\it 3d coordinate}, $\vp_i$, of its center of mass, and a one-hot representation of its 4 properties: {\it color, shape, size}, and {\it material}.
The environment keeps an internal relation graph $G_{adj}$ for all the objects currently in the scene. The relation graph is stored as an adjacency list whose $i^{th}$ entry is a nested array storing $o_i$'s neighbors in 4 cardinal directions {\it left, right, front} and {\it behind}. The criterion for $o_j$ to be the neighbor of $o_i$ in certain direction is if $||\vp_j-\vp_i|| \leq r_{\max}$ and the angle between $\vp_j-\vp_i$ and the cardinal vector of that direction is smaller than $\beta_{\max}$. After every interaction between the agent and the environment, $o_i$ and the relation graph are updated to reflect the current world state.

The agent takes the form of a point mass that can teleport around, which is a mild assumption for standard robotic arms (Other agents are possible as well). Before each interaction, the environment stores a set of language statements that are not satisfied by the current world state. These statements are re-evaluated after the interaction. The statements whose values change to True during the interaction can be used as the goals or instructions for relabeling the trajectories (cf. pre and post conditions used in classical AI planning). Assuming the low-level policy only follows a single instruction at any given instant, the reward for every transition is $1$ if the goal is achieved and $0$ otherwise. The action space we use in this work consists of a point mass agent pushing one object in 1 of the 8 cardinal directions for a fixed number of frames, so the discrete action space has size $8 k_t$, where $k_t \leq 5$ is the number of objects.

\subsection{Tasks}
\label{app:task}

The high-level policy's reward function can be tailored towards the task of interests. We propose three difficult benchmark tasks with extremely sparse rewards.

\subsubsection{Five Object Settings (Standard)}
In this setting, we have a fixed set of 5 spheres of different colors \emph{cyan, purple, green, blue, red}.

The first task we consider is \textbf{object arrangement}. We sample a random set of statements that can be simultaneously satisfied and, at every time step, the agent receives a reward of -10.0 if at least 1 statement is not satisfied satisfied and 0.0 only if all statements are satisfied. At the beginning of every episode, we reset the environment until none of the statements is satisfied. The exact arrangement constraints are:
\textbf{(1)} red ball to the right of purple ball;
\textbf{(2)} green ball to the right of red ball;
\textbf{(3)} green ball to the right of cyan ball;
\textbf{(4)} purple ball to the left of cyan ball;
\textbf{(5)} cyan ball to the right of purple ball;
\textbf{(6)} red ball in front of blue ball; 
\textbf{(7)} red ball to the left of green ball;
\textbf{(8)} green ball in front of blue ball;
\textbf{(9)} purple ball to the left of cyan ball;
\textbf{(10)} blue ball behind the red ball

The second task is \textbf{object ordering}. An example of such a task is {\it``arrange the objects so that their colors range from red to blue in the horizontal direction, and keep the objects close vertically"}. In this case, the configuration can be specified with 4 pair-wise constraint between the objects. We reset the environment until at most 1 pair-wise constraint is satisfied involving the x-coordinate and the y-coordinate. At every time step, the agent receives a reward of -10.0 if at least 1 statement is not satisfied and 0.0 only if all statements are satisfied. The ordering of color is: \emph{cyan, purple, green, blue, red} from left to right.

The third task is \textbf{object sorting}. In  this task, the agent needs to sort 4 object around a central object; furthermore, the 4 objects cannot be too far away from the central object. Once again,  the agent receives a reward of -10.0 if at least 1 constraint is violated, and 0.0 only if all constraints are satisfied. Environment is reset until at most 1 constraint is satisfied. Images of the end goal for each high-level tasks are show in Figure \ref{fig:highlevelgoal}.

\subsubsection{Diverse Object Settings}
Here, instead of 5 fixed objects, we introduce 3 different shapes \emph{cube, sphere} and \emph{cylinder} in combinations with 5 colors. Both colors and shapes can repeat but the same combination of color and shape does not repeat. In this setting, there are $ \binom{15}{5}=3003$ possible object configurations. In this setting, we define the color hierarchy to be \emph{red, green, blue, cyan, purple} from left to right and the shape hierarchy to be \emph{sphere, cube, cylinder} from left to right. Sample goal states of each task are shown in \ref{fig:highlevelgoal}.

The first task is \textbf{color ordering} where the agent needs to manipulate the objects such that their colors are in ascending order.

The second task is \textbf{shape ordering} where the agent needs to manipulate the object such that their shapes are in ascending order.

Finally, the last task is \textbf{color \& shape ordering} where the agent needs to manipulate the object such that the color needs to be in ascending order, and within each color group the shapes are also in ascending order.

Like in the fixed object setting, the agent only receives 0 reward when the objects are completely ordered; otherwise, the reward is always -10.

\vspace{-3mm}
\subsection{Implementation details}
\label{app:impl}

\textbf{Language supervisor}. In this work, each language statement generated by the environment is associated with a {\it functional program} that can be executed on the environment's relation graph to yield an answer that reflects  the value of that statement on the current scene. The functional programs are built from simple elementary operation such as querying the property of objects in the scene, but they can represent a wide range of statements of different nature and can be efficiently executed on the relation graph. This scheme for generating language statements is reminiscent of the CLEVR dataset \cite{clevr} whose code we drew on and modified for our use case. Note that a language statement that can be evaluated is equivalent to a {\it question}, and the instructions we use also take the form of questions. For simplicity and computational efficiency, we use a smaller subset of question family defined in CLEVR that only involves pair-wise relationships (\emph{one-hop}) between the objects. We plan to scale up to full CLEVR complexity and possibly beyond in future works.

\textbf{State based low-level policy.}
When we have access to the ground truth states of the objects in the scene, we use an object-centric representation of states by assuming
$\vs_t = \{o_i\}_{i=1}^{k_t}$,
where  $o_i \in \sR^{d_o}$ is the state representation
of object $i$, and $k_t$ is the number of objects (which can change over time). We also assume $\va_t = \{\alpha_i\}_{i=1}^{k_t}$,
where  each $\alpha_i \in \sR^{d_\alpha}$ acts on individual object $\vo_i$.

\begin{figure}[htbp]
\centering

\includegraphics[width=0.8\textwidth]{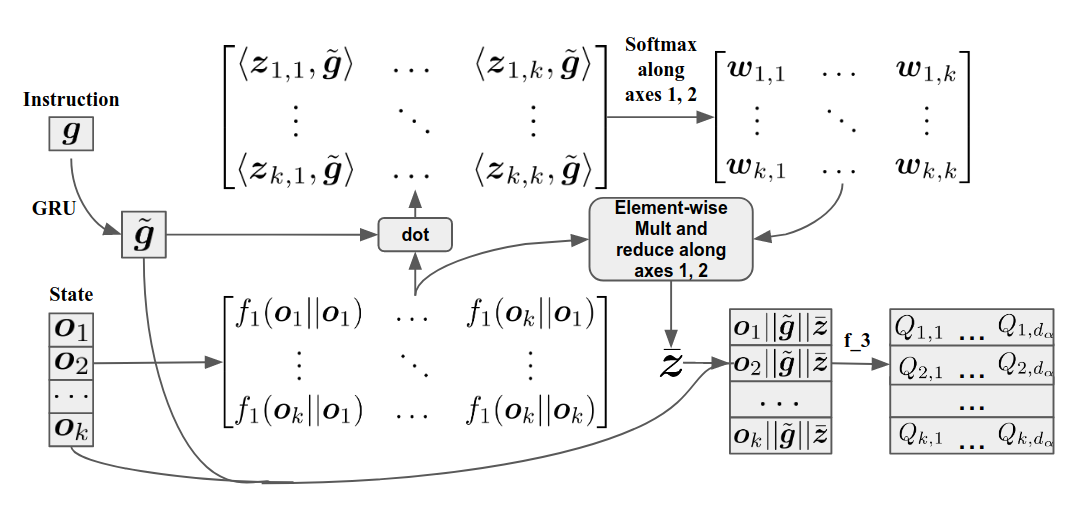}

\caption{Computation graph of the state-based low level policy.}
\label{fig:low_level_architecutre}
\end{figure}

We implement a specialized universal value function approximator \cite{uvfa} for this case. To handle a variable number of  relations between the different objects, and their changing properties, we built a goal-conditioned self attention policy network. Given a set of $k$ object $\{\vo_i\}_{i=1}^k$, we first create pair-wise concatenation of the objects, $\gO= \{ \vo_i \Vert \vo_j\}_{j=1, i=1}^{k, k}$. Then we transform every pair-wise vectors with a single neural network $f_1$
into $\gZ = \{f_1(\vo_i \Vert \vo_j)\}_{j=1,i=1}^{k, k}$. A recurrent neural network with GRU \cite{gru}, $f_2$, embeds the instruction $\vg$ into a real valued vector $\widetilde{\vg} = f_2(\vg)$. We use the embedding to attend over every pair of objects to compute weights $\{w_i = \langle \widetilde{\vg}, \vz_i \rangle | \vz_i \in \gZ\}$.
We then compute a weighted combination of all $p_i$ where the weights are equal to the softmax weights $\exp(w_i)/\sum_{j=1}^{k^2}\exp(w_j)$. 
This combination transforms the elements of $\gZ$ into a single vector $\bar{\vz}$ of fixed size. Each $\vo_i$ is concatenated with $\widetilde{\vg}$ and $\bar{\vz}$ into $\vo_i'=(\vo_i \Vert \widetilde{\vg} \Vert \bar{\vz})$. Then each $\vo_i'$ is transformed with the another neural network $f_3$ whose output is of dimension $d_\alpha$. The final output $Q= \{f_3(\vo_i \Vert \widetilde{\vg} \Vert \bar{\vz})\}_{i=1}^{k}$ is in $\sR^{k \times d_\alpha}$ which represents all state-action values of the state. Illustration of the architecture is shown in Figure \ref{fig:low_level_architecutre}.

\textbf{Image based low-level policy.} In reality, we often do not have access to the state representation of the scene. For many application, a natural alternative is images. We assume $\vs_t \in [0, 1]^{W\times H \times C}$ is the available image representation of the scene (in all experiemnts, W=64, H=64, C=3). Further, we need to adopt a more general action space since we no longer have access to the state representation (e.g. coordinates of the location). To that end, we discretize the 2D space into $10\times10$ grids and an action involves picking an starting location out of the $100$ available grid and a direction out of the the 8 cardinal direction to push. This induces an $800$ dimensional discrete action space.

It is well-known that reinforcement learning from raw image observation is difficult, and even off-policy methods require millions of interaction on Atari games where the discrete action space is small. Increasing the action space would understandably make the already difficult exploration problem harder. A potential remedy is found in the fact that these high-dimensional action space can often be factorized into semantically meaningful groups (e.g. the pushing task can be break down into discrete bins of the $x$ and $y$ axes as well as a pushing direction). Previous works attempted to leverage this observation by using auto-regressive models or assuming conditional independence between the groups \cite{metz2017discrete, tavakoli2018action}. We offer a new approach that aims to make the least assumptions. Following the group assumption, we assume there exists $m=3$ groups
and each group consists of $k_m$ discrete  {\it sub-actions} (i.e. $\gA_m = \{\va_m^{(1)}, \va_m^{(2)}, \dots \va_m^{(k_m)}\}$). Following this definition, we can build a bijective look-up map $\zeta$ between $\gA$ to tuples of sub-actions:
\begin{align}
\gB &= \prod_{n=1}^m\{1, 2, \dots, k_n\} \\
\gA &\overset{\zeta}{\Longrightarrow}\{ (\va_1^{(i_1)}, \dots, \va_m^{(i_m)})\: |\: \forall (i_1, \dots, i_m) \in \gB  \}
\end{align}
We overload the notion $\va_m^{(i)}(\vs)$ to the {\it action feature} of $\va_m^{(i)}$ conditioned on the state and goal and $\zeta(\va,\vs)$ to be a tuple of the corresponding action features. Then the value function each action can be represented as:
\begin{equation}
Q(\vs,\va) = f_\psi(\zeta(\va,\vs))
\end{equation}
where $f_\psi$ is a single neural network parameterized by $\psi$ that is shared between all $\va$. This model does not require picking an order like the auto-regressive model and does not assume conditional independence between the groups. Most importantly, the number of parameters scales sublinearly with the dimension of the actions. The trade-off of this model is that it can be memory-expensive to model the full joint distribution of actions at the same time; however, we found that this model performs empirically well for the pushing tasks considered in this work. We will refer to this operation as \emph{Tensor Concatenation}.

The overall architecture for the UVFA is as follows: the image input is fed through 3 convolution layers with kernel size $\{8, 5, 3\}$, stride $\{2, 2, 1\}$, and channel $\{46, 128, 64\}$; each convlution block is FiLM'd \cite{perez2018film} with the instruction embedding. Then the activation is flattened spatially to $256\times64$ and projected to $28 \times 64$. This is further split into 3 action group of sizes $\{ 10\times64, 10\times64, 8\times64\}$ and fed through tensor concatenation. $f_\psi$ is parameterized by a 2-layer MLP with 512 hidden units at each layer and output dimension of 1 which is the Q-value.

\begin{figure}[htbp]
\centering

\includegraphics[width=0.8\textwidth]{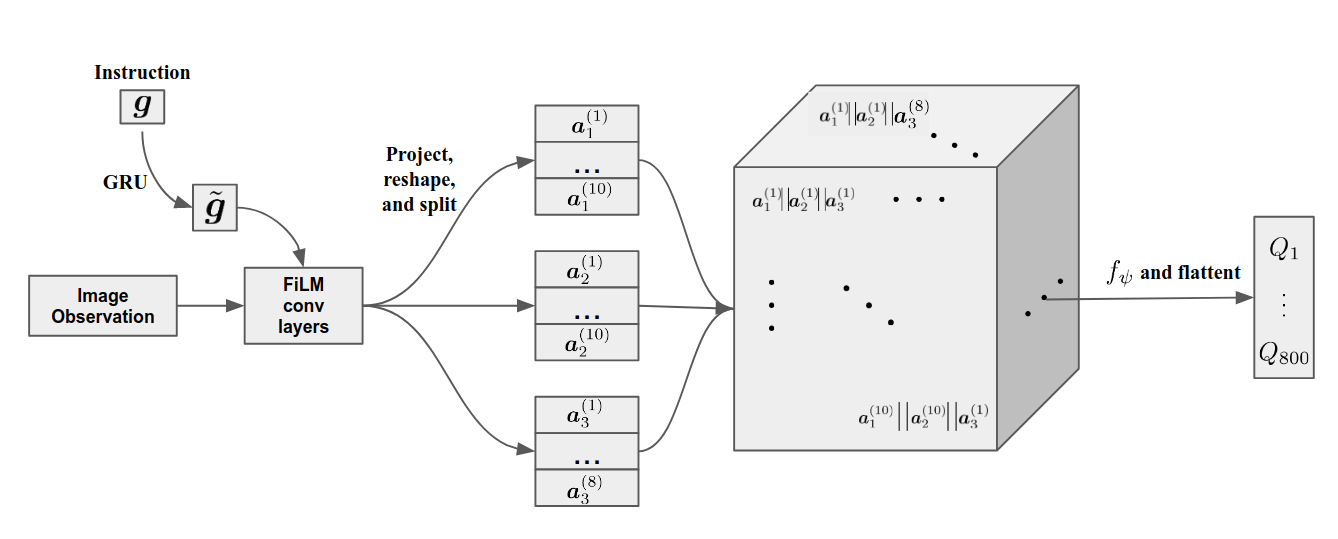}

\caption{Computation graph of the vision-based low level policy.}
\label{fig:image_low_level_architecutre}
\end{figure}

Both policy networks are trained with HIR. Training mini-batches are uniformly sampled from the replay buffer.
Each episode lasts for 100 steps. When the current instruction is accomplished, a new one that is currently not satisfied will be sampled. To accelerate the initial training and increase the diversity of instructions, we put a 10 step time limit on each instruction, so the policy does not get stuck if it is unable to finish the current instruction.

\textbf{High-level policy.} For simplicity, we use a Double DQN \cite{doubleq} to train the high-level policy. We use an instruction set that consists of 80 instructions ($|\gI| = 80$ for standard and $|\gI|=240$ for diverse)
that can sufficiently cover all relationships between the objects. We roll out the low-level policy for 5 steps for every high-level instruction ($T'=5$). Training mini-batches are uniformly sampled from the replay buffer. The state-based high-level policy uses the same observation space as the low-level policy; the image-based high-level policy uses extracted visual features from the low-level policy and then extracts salient spatial points with spatial softmax \cite{levine2016end}. The convolutional weights are frozen during training. This design choice makes natural sense since humans also use a single visual cortex to process all initial visual signals, but training from scratch is certainly possible, if not ideal, should computation budget not matter. For the diverse visual tasks, we found that using the convolutional features with spatial softmax could not yield a good representation for the downstream tasks. Due to time constraints, the experiments shown for the diverse high-level tasks use the ground truth state, namely position, one-hot encoded colors, and shapes for the high-level policy; however, the low-level policy only has access to the image. We believe a learned convolutional layer would solve this problem. Finally, we note that the high-level policy picks each sentence independent and therefore does not leverage the structure of language. While generating language is generally hard, a generative model would have more natural resemblance to \emph{thought}. We think this is an extremely important and exciting direction for future work.

\newpage
\section{Algorithms}

In this section we elaborate on our proposed algorithm and lay out finer details.

\subsection{Overall algorithm}
\label{app:overall_alg}
The overall hierarchical algorithm is as follows:
\begin{algorithm}[htb]
   \caption{Overall Hierarchical Training}
   \label{alg:overall}
\begin{algorithmic}[1]
   \STATE {\bfseries Inputs:} Low level RL algorithm $\mathscr{A}_l$; High-level RL algorithm $\mathscr{A}_h$; Environment $\mathscr{E}$; other relevant inputs of Algorithms \ref{alg:hir} and \ref{app:highlevelalg}
   \STATE $\pi_{l}(\va|\vs, \vg) \leftarrow$ low-level policy trained with $\mathscr{A}_l$ and other appropriate inputs (Algorithm \ref{alg:hir})
   \STATE $\pi_h(\vg|\vs) \leftarrow$ high-level policy trained with $\mathscr{A}_h$, $\pi_{l}(\va|\vs, \vg)$ and other appropriate inputs (Algorithm \ref{app:highlevelalg})
   \STATE {\bfseries return} $\pi_{l}(\va|\vs, \vg)$ and $\pi_{h}(\vg|\vs)$ 
\end{algorithmic}
\end{algorithm}

\subsection{Training the low-level policy}
\label{app:low_alg}
For both state-based and vision-based experiments, we use DDQN as $\mathscr{A}_l$. For the state-based experiments, the agent receives binary reward based on whether the action taken completes the instruction; for the vision-based experiments, we found it instrumental to add an object movements bonus, i.e. if the agents change the position of the objects by some minimum threshold, the agetn receives a 0.25 reward. This alleviates the exploration problem in high dimensional action space ($\sR^{800}$), but the algorithm is capable of learning without the exploration bonus. (Algorithm \ref{alg:hir}). We adopt the similar setting as HER where unit of time consists of epochs, cycles and episode. Each cycle consists of 50 episode and each episode consists of 100 steps. While we set the number of epoch to 50, in practice we never actually reach there. We adopt an epsilon greedy exploration strategy where at the beginning of every cycle we multiply the exploration by a factor of 0.993, starting from 1 but at the beginning we 10 cycles to populate the buffer. The minimum epsilon is 0.1. We use $\gamma=0.9$ and replay buffer of size 2e6. The target network we use is a 0.95 moving average of the model parameters, updated at the beginning of every cycle. Every episode, we update the network for 100 steps with the Adam Optimizer at minibatch of size 128 randomly sampled from the replay buffer.

\begin{algorithm}[htb]
   \caption{RL with hindsight instruction relabeling (HIR)}
   \label{alg:hir}
\begin{algorithmic}[1]
   \STATE {\bfseries Inputs:} off-policy RL algorithm $\mathscr{A}_l$; instruction relabeling strategy $\mathscr{S}$; language supervisor $\Omega$;
   Environment $\mathscr{E}$; number of relabeled future $K$
   \STATE {\bfseries Initialize} replay buffer $\sB$ and $\pi_l(\va|\vs, \vg)$
   \FOR{episode $i=1$ {\bfseries to} $M$}
        \STATE $\vs_0 \leftarrow$ reset $\mathscr{E}$
        \STATE $\vg \sim \gU(\{\vg \in \Omega(\vs_0) \: | \:\Psi(\vs_{0}, \vg)=0\})$
        \STATE $\tau \leftarrow$ [ ]
        \FOR{step $t=0$ {\bfseries to} $T$}
            \STATE $\sU_t \leftarrow \{\vg \in \Omega(\vs_t) \: | \: \Psi(\vs_{t}, \vg)=0\}$
            \STATE $\va_t \sim \pi_{\mathscr{A}}(\va|\vs_t, \vg)$
            \STATE $\vs_{t+1}\leftarrow$ Take action $\va_t$ from $\vs_{t}$
            \STATE $r_{t}\leftarrow$ $\Psi(\vs_{t+1}, \vg)$
            \STATE $\sV_t \leftarrow \sU_t \setminus \{\vg \in \Omega(\vs_{t+1}) \: | \: \Psi(\vs_{t+1}, \vg)=0\}$
            \STATE add $(\vs_{t},\va_t, \vg, r_t, \vs_{t+1},\va_{t+1}, \sV_t)$ to $\tau$
            \IF{$r_t = 1$}
            \STATE $\vg \sim \gU(\{\vg \in \Omega(\vs_{t+1}) \: | \: \Psi(\vs_{t+1}, \vg)=0\})$
            \ENDIF
        \ENDFOR
        \FOR{step $t=0$ {\bfseries to} $T$}
            \STATE $\sB \leftarrow \sB \cup \{(\vs_{t},\va_t, \vg, r_t, \vs_{t+1},\va_{t+1})\}$
            \FOR{$\vg' \in \sV_t$ }
                \STATE $\sB \leftarrow \sB \cup \{(\vs_{t},\va_t, \vg', 1, \vs_{t+1},\va_{t+1})\}$
            \ENDFOR
            \STATE $\sW \leftarrow \mathscr{S}(\tau, t, K)^\dagger$
            \FOR{$(\vg', r') \in \sW$ }
                \STATE $\sB \leftarrow \sB \cup \{\vs_{t},\va_t, \vg', r', \vs_{t+1},\va_{t+1})\}$
            \ENDFOR
        \ENDFOR
-        \STATE Update $\pi_{l}(\va|\vs, \vg)$ with $\mathscr{A}_l$ using minibtach from $\sB$
   \ENDFOR
   \STATE {\bfseries return} $\pi_{l}(\va|\vs, \vg)$ 
   \STATE{{\footnotesize $\dagger$ Details in Appendix \ref{app:relabel}; $\gU(\cdot)$ denotes uniform sampling from the given set.}}
\end{algorithmic}
\end{algorithm}

\subsection{Training the high-level policy}
\label{app:high_alg}
$\mathscr{A}_h$ is also a DDQN. One single set of hyperparameter is used for standard experiments and another for the diverse experiments. DDQN is trained for 2e6 steps, uses uniform replay buffer of size 1e5, linearly anneals epsilon from 1 to 0.05 in 3e5 steps, Adam and batch size 256. $T' = 5$ for all our experiments, but the experience the network sees is equivalent of 1 step. For the diverse settings, due to time constraint, we use priority replay buffer of size 1e6 for all diverse experiment including the DDQN baselines. (Algorithm \ref{app:highlevelalg}) We use the Adam Optimizer \cite{kingma2014adam} with initial learning rate of 0.0001. Discount factor $\gamma=0.9$. Learning starts after 100 episodes each of which lasts for 100 steps (100 high-level actions).

\begin{algorithm}[htb]
  \caption{Training high-level policy}
  \label{app:highlevelalg}
\begin{algorithmic}[1]
  \STATE {\bfseries Inputs:} Any RL algorithm $\mathscr{A}_h$; reward function $R: \gS \rightarrow [r_{\min}, r_{\max}]^*$; instruction set $\gI$; instruction encoder $\phi$; low-level policy $\pi_l(\va|\vs, \vg)$
  \STATE {\bfseries Initialize} $\mathscr{A}$
  \FOR{episode $i=1$ {\bfseries to} $M$}
        \STATE $\vs_0 \leftarrow$ reset $\mathscr{E}$
        \FOR{step $t=0$ {\bfseries to} $T$}
            \STATE $\vg \leftarrow$ Sample from $\gI$ using $\pi_h(\vg| \vs_t)$
            \STATE $\vs' \leftarrow \vs_t$
            \FOR{substep $t'=1$ {\bfseries to} $T'$}
                \STATE $\va' \sim \pi_l(\va|\vs', \vg)$
                \STATE $\vs' \leftarrow$ Take action $\va'$ from $\vs'$
            \ENDFOR
            \STATE $\vs_{t+1} \leftarrow \vs'$
            \STATE{Store experience}
        \ENDFOR
        \STATE Update $\mathscr{A}_h$ accordingly with experience collected
  \ENDFOR
  \STATE{{\footnotesize *Here we assume the reward is only based on the new state for simplicity}}
\end{algorithmic}
\end{algorithm}

\subsection{Relabeling Strategy}
\label{app:relabel}
HER \citep{her} demonstrated that the relabeling strategy for trajectories can have significant impacts on the performance of the policy. The most successful relabeling strategy is the ``k-future'' strategy where the goal state and the reward are relabeled with $k$ states in the trajectories that are reached {\it after} the current time step and the reward is discounted based on the discount factor $\gamma$ and how far away the current state is from the future state in $\ell_2$ distance. We modify this strategy for relabeling a language conditioned policy. One challenge with language instruction is that the notion of distance is not well defined as the instruction is under-determined and only captures a part of the information about the actual state. As such, conventional metrics for describing distance between sequences of tokens (e.g. edit distance) do not actually capture the information we are interested in. Instead, we adopt a more ``greedy'' approach to relabeling by putting more focus on 1-step transition where the instruction is actually fulfilled. Namely, we add all transition tuples in $\sV_t$ to the replay buffer $\sB$ (Algorithm \ref{alg:hir}). For future relabeling, we simply use the reward discounted by time steps into the future to relabel the trajectory. While the discounted reward does not usually capture the ``optimal'' or true discounted reward, we found it provides sufficient learning signal. Detailed steps are shown below (Algorithm \ref{app:relabel_alg}). In our experiments, we use $K=4$.

\begin{algorithm}[htb]
  \caption{Future Instruction Relabeling Strategy ($\mathscr{S}$)}
  \label{app:relabel_alg}
\begin{algorithmic}[1]
  \STATE {\bfseries Inputs:} Trajectory $\tau$; current time step $t$; number of relabeled future $K$
  \STATE $\Delta \leftarrow$  [ ] 
    \STATE{\texttt{count} $\leftarrow 0$}
    \WHILE{$\texttt{count} < K$}
        \STATE \texttt{future} $\sim \gU(\{t+1, \ldots, |\tau|\})$
        \STATE $(\vs,\va, \vg, r, \vs',\va', \sV) \leftarrow \tau[\texttt{future}]$
        \IF{$|\sV| > 0$}
            \STATE $\vg' \sim \gU(\sV)$
            \STATE $r' \leftarrow r \cdot \gamma^{\texttt{future}-t}$
            \STATE Store $(\vg', r')$ in $\Delta$
            \STATE $\texttt{count} \leftarrow \texttt{count}+1$
        \ENDIF
    \ENDWHILE
    \RETURN $\Delta$
\end{algorithmic}
\end{algorithm}

In additon to relabeling, if an object is moved (using 800 dimensional action space), we add to replay buffer a transition where the instruction is the name of the object such as \emph{``large rubber red ball''} with associated reward of $1.0$. We found this helps the agent to learn the concept of objects. We refer to this operation as \emph{Unary Relabeling}.

\section{Experimental Details}
\label{app:exp_detail}

\subsection{One-hot encoded representation}
\label{app:ohe}
We assign each instruction a varying number of bins in the one-hot vector. Concretely, we give each instruction of the 600 instruction 1, 4, 10, and 20 bins in the one-hot vector, which means the effective size of the one-hot vector is 600, 2400, 6000 and 12000. When sampling goals, each goal is uniformly dropped into one of its corresponding bins. The one-hot vector is embedded with a 2 layer MLP with 64 hidden units at each layer.

\subsection{Non-compositional representation}
\label{app:nc}
To faithfully evaluate the importance of compositionality, we want a representation that carries the identical information as the language instruction but without the explicit compositional property (but perhaps still to some degree compositional). To this end, we use a Seq2Seq \cite{seq2seq} autoencoder with 64 hidden units to compress the 600 instructions into real-valued continuous vectors. The original tokens are fully recovered which indicates that the compression is {\it lossless} and the latent's information content is the same as the original instruction. This embedding is used in place of the GRU instruction embedding. We also observed that adding regularization to the autoencoder decreases the performance of the resulting representation. For example, decreasing the bottleneck size leads to worse performance, as does adding dropout. Figure \ref{fig:lowlevel} uses an autoencoder with dropout of 0.5 while \ref{tab:generalization} uses one with no dropout. As shown in the experiments, the performance without dropout is better than the one with. We hypothesize that adding regularization decreases the compositionality of the representation.

\subsection{Bag-of-Words representation}
\label{app:bow}
Bag-of-words is a popular technique in natural language processing where a sentence or corpus of texts are treated as a histogram of tokens -- i.e. the order of the tokens are discarded and only the frequency at which a token occurs is considered. For each instruction, we count the occurrence of each token available in the vocabulary and normalize the count by the length of the instruction. This vector is then embedded with a 2 layer MLP with 64 hidden units at each layer.

\subsection{Non-hierarchical baseline}
We use the Double DQN implementation from OpenAI baselines\footnote{https://github.com/openai/baselines/tree/master/baselines/deepq}. We use a 2 layer MLP with 512 hidden units at each layer with the same action dimension as the low-level policy.

\subsection{HRL baselines}
\label{app:other_hrl}
In general, we note that it is difficult to compare different HRL algorithms in an apples-to-apples manner.
HIRO assumes a continuous goal space so we modified the goal to be an $\sR^{10}$ vector representing the locations of each object rather than language. In this regime, we observed HIRO was unable to make good progress. We hypothesize that the highly sparse reward might be the culprit. It is also worth noting that HIRO uses a goal space in $\sR^2$ for navigation (which is by itself a choice of abstraction because the actual agent state space is much higher) while ours is of higher dimensionality. The Tensorflow implementation of HIRO we used can be found at the official Tensorflow repository\footnote{https://github.com/tensorflow/models/tree/master/research/efficient-hrl}. (This is the implementation from the original author).

Option-critic aims to learn everything in a complete end-to-end manner which means it does not use the supervision from language. It is unsurprising that the sparse tasks do not provide sufficient signal for OC. In other words, our method enjoys the benefit of a flexible but fixed abstraction while OC needs to learn such abstraction.
We tried 8, 16, and 32 options for OC but our method has much more sub-policies due to the combinatorial nature of language. The OC implementation in Tensorflow we used is open-sourced\footnote{https://github.com/yadrimz/option-critic}.

\subsection{Hardware Specs and Training time}
All of our experiments are performed on a single Nvidia Tesla V100. We are unable to verify the specs of the virtual CPU. The low-level policy for state-based observation takes about 2 days to train and, for image-based observation, 6 days. The high-level policies for the state-based observation takes about 2 days to train and 3 days for image-based observations (wall-clock time). The implementations are not deliberately optimized for performance as the major bottleneck is actually the language supervisor so it is very likely the time could be dramatically shortened.

\section{More Experimental Results and Discussions}
\label{app:analysis}

\subsection{Low-level policy for a diverse environment}
\begin{figure}[!htbp]
\centering
\centering
\includegraphics[width=0.4\textwidth]{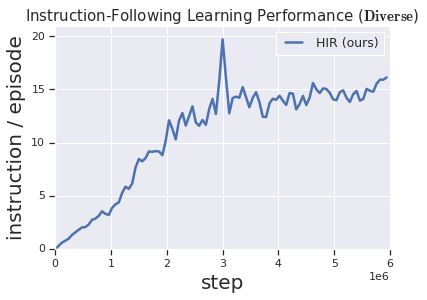}

\caption{\footnotesize Results training the low-level policy on the diverse environment.}
\label{fig:diverse_hir}
\vspace{-0.3cm}
\end{figure}

Figure \ref{fig:diverse_hir} shows the training instruction per episode on the diverse environment. We see that the performance is worse than a fixed number of objects with the same amount of experience. This is perhaps not surprising considering the visual tasks are much more diverse and hence more challenging.

\subsection{Proposed environment, sparse reward, and structured exploration}
While DDQN worked on 2 cases in the state-based environment, it is unable to solve any of the problems in visual domain. We hypothesize that the pixel observations and increase in action space ($20\times$ increase) makes the exploration difficult for DDQN. The difficulty of the tasks -- in particular the 3 standard tasks -- is reflected in the fact that the reward from non-hierarchical random action is stably 0 with small variance, meaning that under the sparse reward setting the agent almost never visits the goal states (states with non-zero reward). On the other hand, the random exploration reward is much higher for our method as the exploration in the space of language is \emph{structured} as the exploration is aware of the task-relevant structure of the environment.

\end{document}